%% file: main.tex
\documentclass[11pt, letterpaper]{nyu}
\usepackage[all]{hypcap}

\usepackage{hyperref}[citecolor=lightblue]

\hypersetup{
    colorlinks = true,
}

\usepackage{microtype}
\usepackage{graphicx}
\usepackage{subfigure}
\usepackage{booktabs} 
\usepackage{float}
\usepackage{amsmath}
\usepackage{amssymb}
\usepackage{mathtools}
\usepackage{amsthm}
\usepackage{caption}
\usepackage{mathrsfs}
\usepackage{nicefrac}
\usepackage{dsfont}
\usepackage{enumitem}
\usepackage{float}
\usepackage[authoryear, sort&compress, round]{natbib}
\usepackage{xspace}
\usepackage[capitalize,noabbrev]{cleveref}
\usepackage{subcaption}
\usepackage{wrapfig}
\usepackage{lipsum}
\usepackage{listings}
\usepackage{amsmath}
\usepackage{amssymb}
\usepackage{mathtools}
\usepackage{amsthm}
\usepackage{bbm}
\usepackage{algpseudocode}
\usepackage{setspace}
\usepackage{color}
\usepackage[skins,theorems]{tcolorbox}
\usepackage{xcolor}
\usepackage{tikz}

\definecolor{deepblue}{rgb}{0,0,0.5}
\definecolor{deepred}{rgb}{0.6,0,0}
\definecolor{deepgreen}{rgb}{0,0.5,0}

\newcommand\pythonstyle{\lstset{
basicstyle=\ttfamily\footnotesize,
language=Python,
morekeywords={self, clip, exp, mse_loss, uniform_sample, concatenate, logsumexp},              
keywordstyle=\color{deepblue},
emph={MyClass,__init__},          
emphstyle=\color{deepred},    
stringstyle=\color{deepgreen},
frame=single,                         
showstringspaces=false
}}

\lstnewenvironment{python}[1][]
{
\pythonstyle
\lstset{#1}
}
{}

\newcommand\pythoninline[1]{{\pythonstyle\lstinline!#1!}}

\usepackage[textsize=tiny]{todonotes}

\usepackage{crossreftools}
\usepackage{dsfont}
\usepackage{nicefrac}
\usepackage{inconsolata}
\usepackage{algorithm}
\usepackage{amssymb}
\usepackage{csquotes}

\usepackage{wrapfig}

\newcommand{\xxnote}[3]{}
\renewcommand{\xxnote}[3]{}

\newcommand{\website}{\href{http://www.e-flesh.com}{e-flesh.com}}

\definecolor{olivegreen}{HTML}{3C8031}

\newcommand{\name}{eFlesh}

\title{\name{}: Highly customizable Magnetic Touch Sensing using Cut-Cell Microstructures}
\websitedef{\website}

\reportnumber{} 

\author{Venkatesh Pattabiraman}
\author{Zizhou Huang}
\author{Daniele Panozzo}
\author{Denis Zorin}
\author{Lerrel Pinto}
\author{Raunaq Bhirangi}

\affil{New York University}

\correspondingauthors={%
  \href{mailto:venkatesh.p@nyu.edu}{venkatesh.p@nyu.edu},\,
  \href{mailto:raunaqbhirangi@nyu.edu}{raunaqbhirangi@nyu.edu}.}

\begin{abstract}
\vspace{-1em}
If human experience is any guide, operating effectively in unstructured environments—like homes and offices—requires robots to sense the forces during physical interaction. Yet, the lack of a versatile, accessible, and easily customizable tactile sensor has led to fragmented, sensor-specific solutions in robotic manipulation—and in many cases, to force-unaware, sensorless approaches. With \name{}, we bridge this gap by introducing a magnetic tactile sensor that is low-cost, easy to fabricate, and highly customizable. Building an \name{} sensor requires only four components: a hobbyist 3D printer, off-the-shelf magnets ($<$ \$5), a CAD model of the desired shape, and a magnetometer circuit board. The sensor is constructed from tiled, parameterized microstructures, which allow for tuning the sensor’s geometry and its mechanical response. We provide an open-source design tool that converts convex OBJ/STL files into 3D-printable STLs for fabrication. This modular design framework enables users to create application-specific sensors, and to adjust sensitivity depending on the task. Our sensor characterization experiments demonstrate the capabilities of \name{}: contact localization RMSE of 0.5 mm, and force prediction RMSE of 0.27 N for normal force and 0.12 N for shear force. We also present a learned slip detection model that generalizes to unseen objects with 95\% accuracy, and visuotactile control policies that improve manipulation performance by 40\% over vision-only baselines -- achieving $91\%$ average success rate for four precise tasks that require sub-mm accuracy for successful completion. All design files, code and the CAD-to-\name{} STL conversion tool are open-sourced and available on \website.
\end{abstract}

\begin{document}
\maketitle

\input{sections/1_intro}
\input{sections/2_results}
\input{sections/3_discussion}
\input{sections/4_materials}


\section*{Acknowledgments}
We thank Anya Zorin, Enes Erciyes, Haritheja Etukuru, Irmak Guzey, Mahi Shafiullah, Siddhant Haldar and Zichen Cui for valuable feedback and discussions.
\paragraph*{Funding:}
This work was supported by grants from Honda, Microsoft, Hyundai, NSF award 2339096, and ONR award N00014-22-1-2773. LP is supported by the Sloan and Packard Fellowships.
\paragraph*{Author contributions:}
V.P. led the project, fabricated the prototypes and conducted experiments, with guidance from L.P. and R.B. on experiment design and project directions. L.P. and V.P. conceived the project. R.B. led the writing of the manuscript, with help from V.P.. Z.H., D.P. and D.Z. provided expertise on microstructures. All authors provided feedback and revisions.
\paragraph*{Competing interests:}
There are no competing interests to declare.
\paragraph*{Data and materials availability:}
All data and materials needed to reproduce and evaluate the results and conclusions of this manuscript are present in the paper or in the Supplementary Materials.

\newpage
\bibliography{references}

\newpage
\appendix
\input{sections/6_appendix}

\end{document}

%% file: sections/1_intro.tex
\begin{figure}[h]
    \vskip -0.5em
    \centering
    \includegraphics[width=0.99\linewidth]{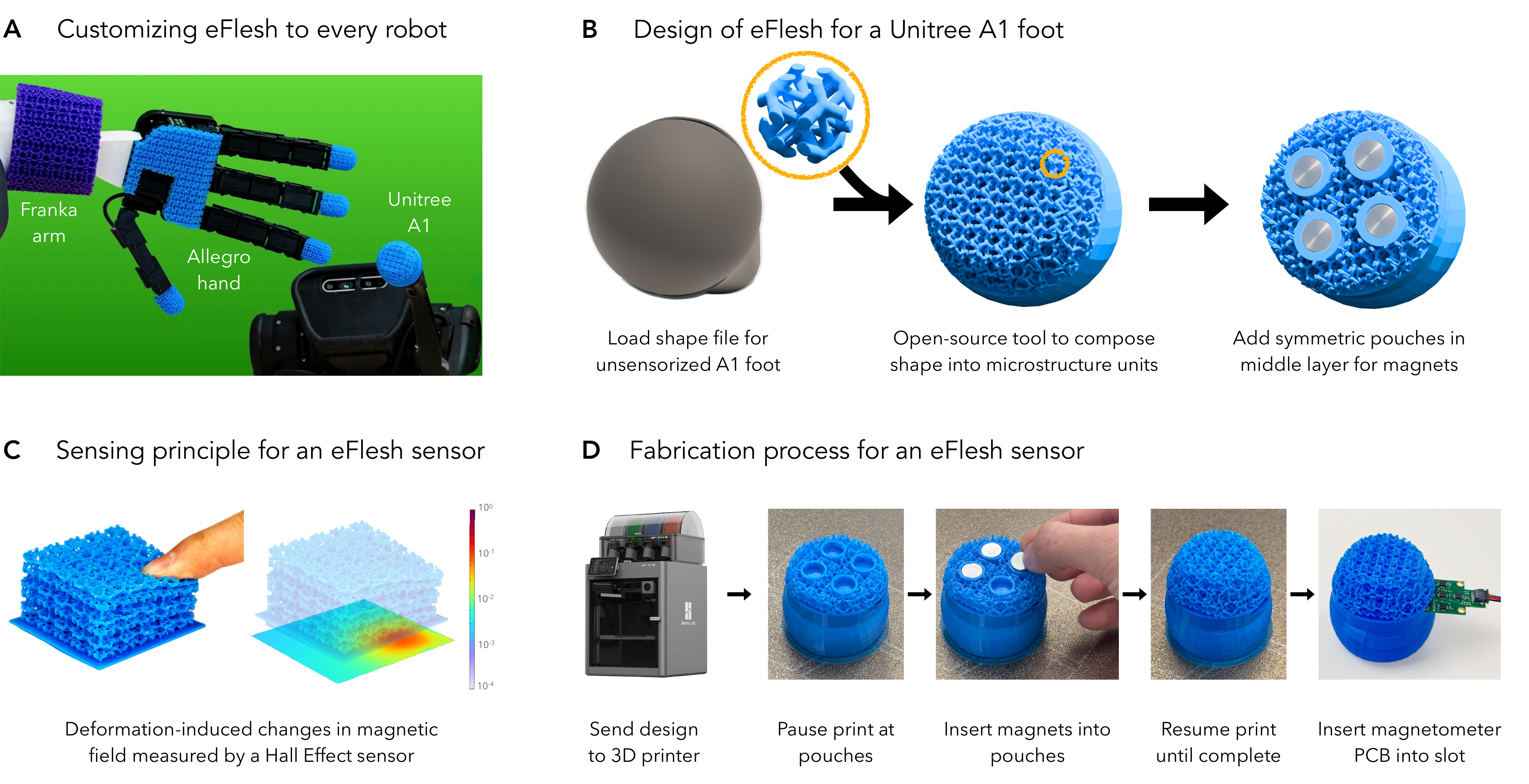}
    \caption{
    \textbf{\name{} is Customizable and Easy-to-Fabricate.}
    (\textbf{A}) \name{} sensors customized and attached to robots of diverse form factors, ranging from hand fingertips to arm sleeves.
    (\textbf{B}) Design overview of our open-source tool, to generate a printable \name{} sensor adapted to a given input shape, herein the Unitree A1 foot.
    (\textbf{C}) Visualization of the change in magnetic flux density, caused by the deformation from a human touching \name{}.
    (\textbf{D}) Fabrication workflow: The print pauses at the completion of the press-fit pouches; a human inserts magnets and resumes the print; finally, a magnetometer circuit board is inserted into the slot to sense magnetic fields.
    }
    \label{fig:fig-1}
\end{figure}
\section{Introduction}
\noindent Physical interaction with the world is nearly unimaginable without our sense of touch. From tying shoelaces or opening a bag of chips to wiping tables or cutting fruits, local force feedback is central to our ability to manipulate objects in our environment. While roboticists have long recognized the necessity of exteroceptive tactile feedback for complex manipulation~\cite{Calandra2018MoreTA,yuan2024robot,yin2024learning, pattabiraman2024learning,huang3d}, most recent work has either entirely forgone the use of tactile sensing~\cite{Fu_2022_CVPR,padalkar2023open, chi2024diffusionpolicy, bharadhwaj2023roboagent, chi2024universal,etukuru2024robot,khazatsky2024droid,lin2025data,xie2024decomposing}, or been restricted to silo-ed, sensor-specific efforts that struggle to command widespread adoption~\cite{zhao2024transferabletactiletransformersrepresentation, bhirangi2024hierarchical}. Furthermore, as deep learning has revolutionized visual and linguistic reasoning, the lack of similar, mutually compatible tactile data has precluded analogous developments in large-scale tactile reasoning. We posit that the lack of a sufficiently versatile, accessible tactile sensor is the primary contributor to these disparities. With \name{}, we seek to tackle this problem by leveraging recent advances in computer graphics and additive manufacturing to create a low-cost, customizable sensor that anyone can fabricate with just a hobbyist 3D printer, off-the-shelf magnets, and less than 5 minutes of active involvement.

Despite the rich existing landscape of tactile sensors across a range of modalities, prior work has struggled to strike the right balance between cost, accessibility and versatility. While there has been some work on rigid tactile sensors~\cite{chengrigid, Fleer2019LearningEH, Zhang2022FingerinspiredRHA}, the need for conformal interfaces for stable contact has placed this field squarely in the soft robotics realm~\cite{hellebrekers2019soft, Yan2021SoftMS,shih2020electronic,kim2020heterogeneous}. Primary among soft sensing technologies are resistive and piezoresistive technologies that rely on strain-based changes in resistance to detect, localize and measure force~\cite{stassi2014flexible, huang3d, sundaram2019learning, pannen2021lowcosteasytomanufactureflexiblemultitaxel, egli2024sensorizedsoftskindexterous}. These sensors are generally low-cost and admit flexible form factors, but are also bulky and failure-prone due to a large number of wires, cannot sense shear forces and offer limited force resolution. Capacitive technologies~\cite{glauser2019deformation, wu2020capacitivo} address the shear force and sensitivity limitations, but remain fragile due to direct connections between circuitry and elastomer~\cite{Xu2015StretchNF, Xu2024CushSenseSS}. Furthermore, they often involve intricate, boutique fabrication methods resulting in significant variability across sensor instances. Another interesting class of sensors is MEMS-based technologies~\cite{wettels2008biomimetic} that combine several sensors like IMU and vibration sensing to create multimodal sensory solutions. However, the use of expensive components and the complexity of fabrication make these sensors impractical for general as well as research use. Another broadly used category of sensors is camera-based optical sensors~\cite{yuan2017gelsight, gelsightmini2023, lambeta2020digit, wang2021gelsight} that use cameras to capture the deformation of a silicone gel and offer high-resolution contact information in addition to detecting 3-axis forces acting on the gel. However, these sensors suffer from bulky, rigid form factors resulting from the physical limits on the camera focal length. Furthermore, they do not admit themselves to arbitrary form factors -- a conspicuous disadvantage highlighted by the complexity in developing optical sensors of different shapes~\cite{liu2023gelsightendoflexsoftendoskeleton, wang2021gelsight, taylor2022gelslim}. Moreover, a pervasive challenge across all of these approaches is the high instance-to-instance variability in sensor response~\cite{lambeta2020digit, suresh2023neural}, which hinders sensor interchangeability and poses a major obstacle to large-scale data collection and deployment.

To overcome these limitations, we turn to magnetic sensing as a promising alternative. Magnetic sensors~\cite{hellebrekers2019soft, bhirangi2021reskin, bhirangi2024anyskin, tomo2018new} are low cost, admit flexible form factors~\cite{bhirangi2021reskin,bhirangi2024anyskin}, multi-axis force sensing, high contact resolution as well as a unique consistency in response across different sensor instances~\cite{bhirangi2024anyskin}. However, this class of sensors suffers from two main drawbacks: magnetic interference and expensive fabrication. With \name{}, we present a magnetic sensor that preserves the advantages of prior magnetic sensors while significantly simplifying the fabrication process, and enhancing customizability of the sensor in terms of form factor as well as sensor response. A core idea that makes this possible is \name{}'s tiled microstructure-based construction.

Recent advances in additive manufacturing have made it significantly easier to fabricate complex geometric structures with user-defined effective material properties. This has opened up customizable design workflows across a range of domains, from strong, lightweight aerospace components to personalized shoe soles, prosthetic devices, and compliant robotic parts~\cite{ntopology, carbon}. A common approach in these applications is to design components as compositions of geometric microstructures from predefined families to achieve desired material response~\cite{Panetta2015, Panetta2017, Zhu2017}. These microstructure families are typically optimized to map effective material properties to printable geometric tiles. Designers can populate an object with these tiles to approximate target mechanical behaviors, without the need for computationally intensive topology or shape optimization. This strategy significantly reduces design overhead and allows for reuse across multiple objects, making it well-suited for rapid, low-cost prototyping. Building on this foundation, we introduce \name{}, a novel magnetic tactile sensing platform that fuses state-of-the-art approaches to parameterized microstructural design with embedded magnetic sensing. Our approach leverages the 3D boundary cell families proposed by Tozoni et al.\cite{Tozoni2024}, extending their utility beyond passive mechanics to active, multipurpose tactile sensing. Arbitrary surface geometries can be sensorized as composite microstructured tiles, as illustrated in Fig.~\ref{fig:fig-1}A, while magnetometer circuit boards are embedded within or beneath the structure to transduce deformation into rich magnetic signals (Fig.~\ref{fig:fig-1}C). This combination of material programmability, scalability, and ease of fabrication positions \name{} as a powerful, versatile solution for integrating tactile feedback into custom robotic systems. In this paper, we present a concrete methodology for the design and fabrication of \name{}, and demonstrate its versatility across a range of robotic tasks and sensing applications. Our main findings can be summarized as follows:

\begin{enumerate}
    \item We present \name{}, a magnetic tactile sensor that can be fabricated into any 3D-printable convex shape with off-the-shelf magnets and Hall sensors.
    \item We characterize the response of \name{} and demonstrate a contact localization RMS error of 0.5 mm, and force prediction RMS errors of 0.27 N for normal force and 0.12 N for shear force (Sections~\ref{subsec:contact-localization},\ref{subsec:normal-force},\ref{subsec:shear-force}).
    \item We integrate \name{} with learning based approaches for slip detection, where we achieve a success rate of 95\% on unseen objects, and visuotactile learning, where we outperform vision-only baselines by 41\%, while achieving $>90\%$ success rates on set of contact-rich, precise manipulation tasks requiring sub-mm precision for success (Sections~\ref{subsec:slip-detection},\ref{subsec:policy-learning}).
    \item We propose a novel technique to circumvent the interference problem that has plagued magnetic sensing, by placing magnets of alternating polarities next to each other, and demonstrate that with this technique, inter-sensor as well as external magnetic interference correspond to signal from less than 1mm of sensor deformation (Section~\ref{subsec:interference}).
\end{enumerate}

All design files, code, and the CAD-to-\name{} STL conversion tool are open-sourced and can be found on \website.

%% file: sections/2_results.tex
\section{Results}
\subsection{A magnetic tactile sensor based on cut cell microstructures}

The core principle underlying a range of magnetic tactile sensors is the coupling of mechanical deformation with perturbations in a magnetic field. These perturbations can be measured using magnetometers~\cite{hall1879new} and reasoned over for a number of contact reasoning tasks such as material identification, object classification, and precise manipulation~\cite{li2018slip, baishya2016robust, lin2019learning, pattabiraman2024learning}. Most prior designs achieve the deformation-magnetic field coupling through the use of magnetic elastomers. For instance, ReSkin~\cite{bhirangi2021reskin} and AnySkin~\cite{bhirangi2024anyskin} embed magnetic microparticles into elastomers and magnetize them using either permanent magnets~\cite{bhirangi2021reskin} or pulse magnetizers~\cite{bhirangi2023all, bhirangi2024anyskin}. Similarly, uSkin~\cite{tomo2018new} incorporates macro-sized permanent magnets within an elastomer matrix to produce magnetic signals under deformation.

Despite their success, these approaches have notable scalability and performance limitations. Microparticle based sensors exhibit high instance-to-instance variability~\cite{tomo2018new, bhirangi2021reskin, bhirangi2023all}, and often require expensive or specialized magnetization procedures~\cite{bhirangi2023all, bhirangi2024anyskin}. Furthermore, the relatively weak magnetic fields from magnetic elastomers constrain the extent of separation between the sensing surface and magnetometer, while also making them vulnerable to cross-talk and environmental interference. Similarly, sensors that embed rigid permanent magnets~\cite{tomo2018new} tend to compromise mechanical durability and sensitivity due to stiff-soft material interfaces, and further hinder scalability through complex, expensive fabrication requirements~\cite{signor2022grad,tomo2018new}. Most importantly, while robot interfaces come in all shapes and forms, most existing sensors require significant domain knowledge and manufacturing expertise for application-specific customizability.

The recent surge in technology and accessibility of rapid prototyping has resulted in a number of works seeking to leverage these advances to create customizable sensors~\cite{zhu3ddeform, guo3dstretch, Vatani2015Combined3P, Zhou2021ThreedimensionalPO, Li20183DPS, Pei2021AF3, He20223DPrintedSS, shih2019design}. However, most of these sensors require intermediate fabrication steps that require significant expertise -- from assembling silver nanoparticles and nanowires~\cite{guo3dstretch} to hydrogel ink design and fabrication~\cite{Li20183DPS}. Other solutions offering shape customizability exhibit high variability across sensor instances~\cite{shih2019design}, while material customizability often requires complex, expensive equipment like air pumps and reservoirs~\cite{He20223DPrintedSS}.

To address these limitations, \name{} is a low-cost, modular, and accessible magnetic tactile sensing platform designed for both customizability and scalability. Our approach leverages parameterized, durable microstructures to decouple sensor design from complex fabrication constraints. The fabrication is fully compatible with consumer-grade 3D printers, lowering the barrier to entry for non-expert users. In addition to the ease of fabrication, we employ machine learning to demonstrate the richness and robustness of \name{} signal. Our experiments also show that \name{} maintains high signal consistency across instances -- enabling scalable deployment scenarios like contact localization, force estimation, and generalizable policy learning.

To accommodate diverse sensor geometries, we adopt a variation of the infill strategy proposed in \cite{Tozoni2024}. Given an arbitrary convex sensor shape, we first generate a voxelized lattice composed of microstructure units, trimmed to fit the desired form. Next, based on the number and size of magnets to embed, we generate symmetric, press-fit pouches in the sensor’s mid-plane to hold them. This modular design framework greatly simplifies the process of sensorizing a variety of surfaces, as illustrated in Fig.~\ref{fig:fig-1}(B and D).

In the remainder of this section, we present a number of experiments demonstrating the capabilities of \name{} across a range of contact reasoning tasks. We start with controlled contact localization and force estimation tasks where we train machine learning models to demonstrate the richness and utility of the \name{} signal. Following this, we explore two deployment-driven scenarios:
\begin{enumerate}
    \item \textbf{Learned slip detection:} A Hello Robot Stretch uses \name{} to predict when a grasped object is being pulled out of its grasp -- demonstrating the sensor’s slip detection capability and deployability in real-time interaction scenarios.
    \item \textbf{Contact-rich visuotactile policy learning:} Based on the VisuoSkin framework presented in~\cite{pattabiraman2024learning}, we integrate \name{} into a policy learning pipeline for four contact-rich tasks: USB insertion, plug insertion, credit card swiping and whiteboard erasing. These tasks highlight the sensor’s utility in capturing fine-grained contact information critical for precise, closed-loop manipulation.
\end{enumerate}

\begin{wrapfigure}{r}{0.3\textwidth}
    \centering
    \vspace{-10pt}
    \includegraphics[width=0.33\textwidth]{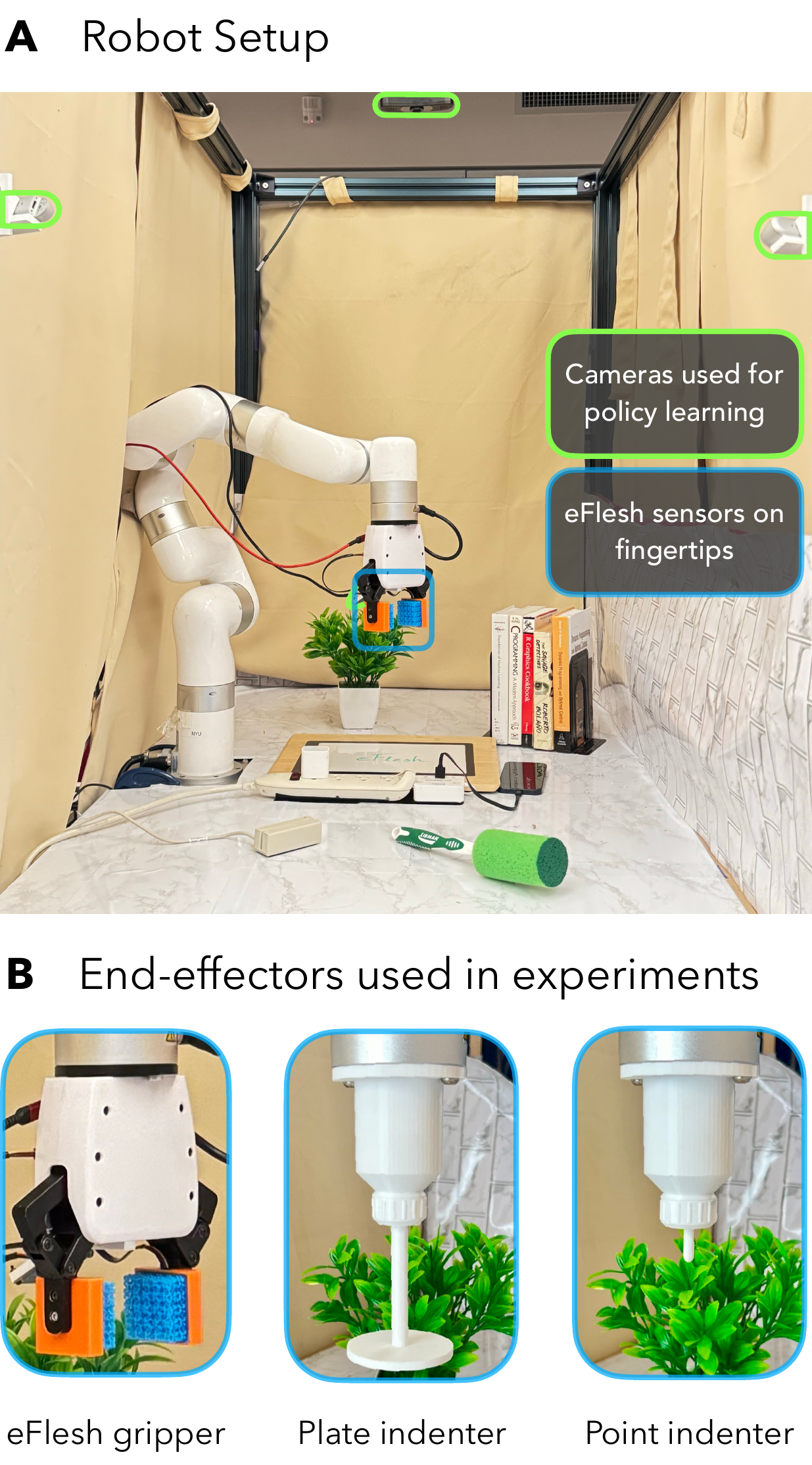}
    \vspace{-10pt}
    \caption{
    \textbf{Robot experiment setup.}
    (\textbf{A}) Our UFactory xArm 7 robot and experimental environment. 
    (\textbf{B}) End-effectors used in policy learning and sensor experiments.
    }
    \label{fig:fig-2}
\end{wrapfigure}

We then analyze the effects of customization on sensor response and resulting operational parameters. Unless stated otherwise, all experiments are conducted using a standardized 40mm x 40mm cuboidal instance of \name{}. Our experimental setup consists of a UFactory xArm 7 robot shown in Fig.~\ref{fig:fig-2}A, equipped with task-specific end effectors (Fig.~\ref{fig:fig-2}B).


\subsection{Contact localization}
\label{subsec:contact-localization}
\begin{figure}[h]
    \centering
    \includegraphics[width=\linewidth]{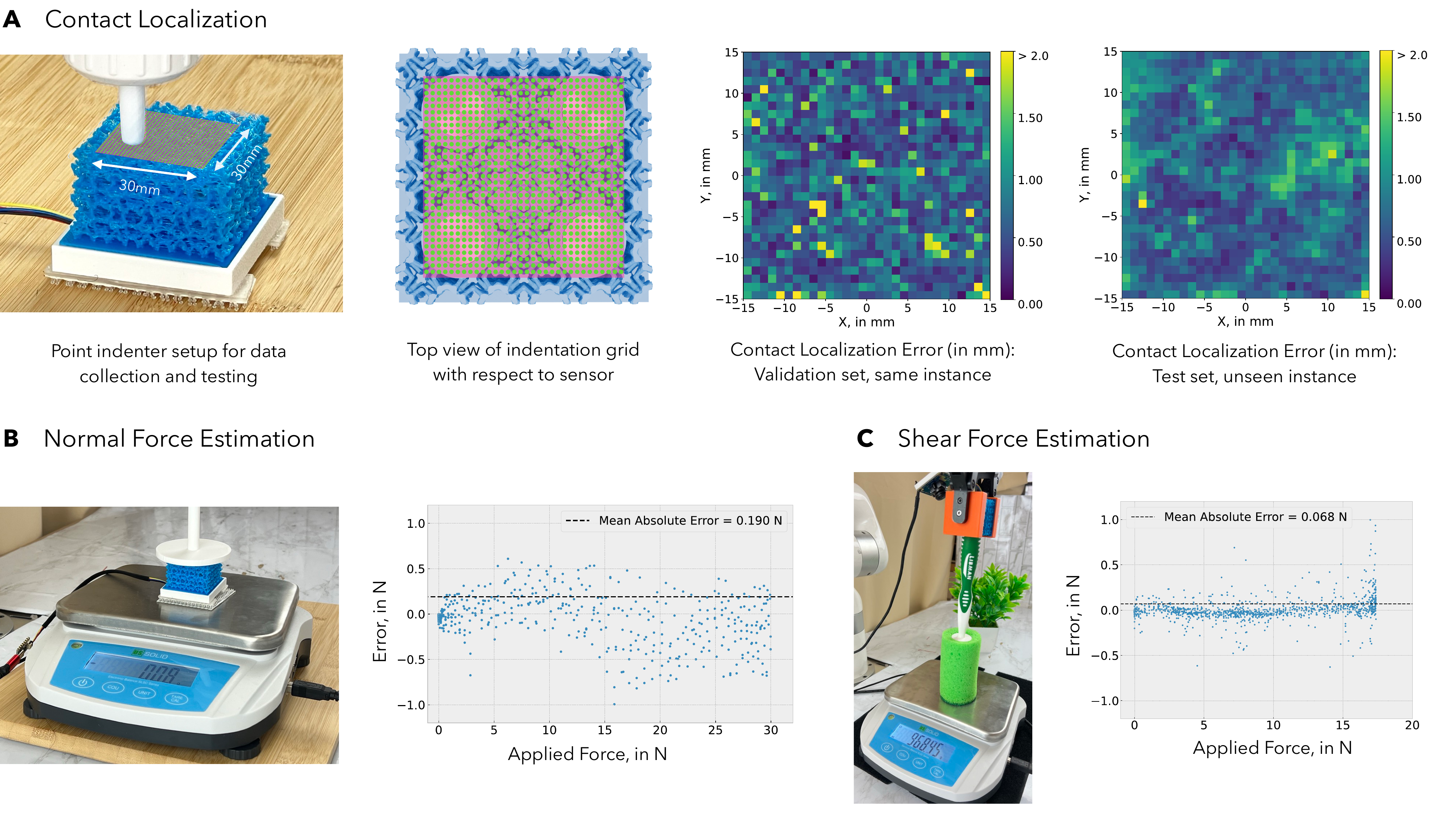}
    \caption{
    \textbf{\name{} Sensor Characterization Experiments.}
    (\textbf{A}), (Left to Right) The UFactory xArm robot arm equipped with a point indenter end-effector probing an \name{} sensor instance for contact localization; Top view of the 30mm $\times$ 30mm indentation grid overlaid on the sensor; Heatmap visualizations of contact localization errors on the validation set of train instance and an unseen \name{} instance.
    (\textbf{B}), xArm with a plate indenter compressing an \name{} sensor on a weighing scale for normal force estimation, alongside the resulting error plot of our trained estimation model.
    (\textbf{C}), xArm equipped with an \name{} gripper pressing a grasped foam stick against a weighing scale, alongside the resulting shear force estimation error plot.
    }
    \label{fig:fig-3}
\end{figure}

An essential factor in selecting a tactile sensor is its contact resolution -- its ability to distinguish between spatially proximal contact points. To evaluate the contact resolution of \name{}, we train a neural network to predict the planar contact location $(x,y)$ relative to the sensor’s center, as well as the contact depth, $z$, relative to its surface, from the raw magnetic field measured by the magnetometers. We place the sensor on a flat table and use a 6mm-diameter 3D-printed hemispherical indenter mounted on a UFactory xArm 7 robot to probe the sensor surface at 1 mm intervals along $x$ and $y$ within a 30mm $\times$ 30mm grid as shown in Fig.~\ref{fig:fig-3}A. For each indentation, we uniformly sample the indentation depth, $z$, between 0.2 mm and 4.2 mm from the sensor surface, and record the change in magnetic field along with the corresponding contact location label, $(x,y,z)$, obtained from robot proprioception. We repeat this process for five passes over the entire grid resulting in a total of 4,500 labeled samples.

Our neural network is a simple MLP with two hidden layers with 128 nodes each and ReLU activations, mapping the 15-dimensional change in magnetic field ($B$) to the 3-dimensional contact location: $(x,y,z)$. We deliberately avoid random sampling when splitting the dataset for training and validation. Instead, we use data from the first four passes over the indentation grid for training, and reserve the fifth pass for validation. This setup better reflects real-world deployment conditions where sensor response tends to drift over time -- a well-documented challenge in soft sensing systems~\cite{hellebrekers2019soft,bhirangi2021reskin}. Under this protocol, the model achieves an \(\mathrm{RMSE}_{x,y}\) of \(0.5\,\mathrm{mm}\) and an \(\mathrm{RMSE}_{\mathrm{depth}}\) of \(0.16\,\mathrm{mm}\) on the validation set, demonstrating that \name{} can localize contact with sub-mm precision. We find that this is slightly worse than a random train-validation split which results in a validation \(\mathrm{RMSE}_{x,y}\) of \(0.4\,\mathrm{mm}\) and a validation \(\mathrm{RMSE}_{\mathrm{depth}}\) of \(0.12\,\mathrm{mm}\), justifying our temporal train-validation split. To better contextualize these results, Fig.~\ref{fig:fig-3}A also visualizes contact localization error at every point of the indentation grid for the validation set, as well as the test set comprised of indentation data from an unseen \name{} instance. These plots illustrate the consistency of low contact localization error over the entire surface of the sensor as well as the generalization of this result to unseen instances of \name{}. 

\subsection{Normal Force estimation}
\label{subsec:normal-force}
To further characterize the force response of \name{}, we train a separate neural network to predict normal force from raw magnetic field signals. We swap the hemispherical point indenter for a flat plate indenter shown in Fig.~\ref{fig:fig-2}B and place a weighing scale underneath the sensor (Fig.~\ref{fig:fig-3}B). To collect data for the experiment, we apply compressive forces ranging from no force to 30 N while recording synchronized measurements from both the sensor and a weighing scale. For training and validation, we adopt a temporal split of the dataset mirroring the protocol used in the contact localization experiment: the first 7,200 data points are used for training, while the remaining 1,800 points form the validation set. We find a normal force prediction RMSE of $0.27$ N, corresponding to a pressure of $125$ Pa, validating \name{}'s ability to capture normal force/pressure with a high degree of precision. Fig.~\ref{fig:fig-3}B also shows the variation of force prediction error as the amount of applied force is increased, illustrating that maximum error remains under 1N across the entire range of applied forces.

\subsection{Shear force estimation}
\label{subsec:shear-force}
A key aspect of human dexterity is the ability to detect impending slip and adjust manipulation strategies accordingly~\cite{bhirangi2024anyskin}. Tactile sensors can enable the same capability in robots if they can capture shear forces acting on the surface of the sensor before an object in contact slips out of grasp. In this experiment, we evaluate \name{}'s capability to estimate planar (shear) forces applied to the sensor surface. To do so, we use a foam cleaning stick (Fig.~\ref{fig:fig-3}C), which enables the application of a broader range of shear forces to the sensor surface without causing slip. The stick is grasped using a parallel jaw gripper, with both gripper tips equipped using \name{}. We then generate data by randomly sampling vertical displacements of the foam stick, pressing it against a weighing scale. During this process, we record synchronized raw magnetic field signals from the sensors and ground-truth force measurements from the scale. The applied shear forces span a range from 0 N to 17.5 N. We train a neural network to predict the planar shear force from the raw magnetic data, using the same temporal split strategy as in previous experiments to account for sensor drift: the first portion of the dataset is used for training, and the remainder is held out for testing. The trained model achieves a root mean squared error (RMSE) of 0.12 N, demonstrating \name{}’s effectiveness in accurately estimating shear forces.

\subsection{Slip detection using deep learning}
\label{subsec:slip-detection}
\begin{figure}
    \centering
    \includegraphics[width=\linewidth]{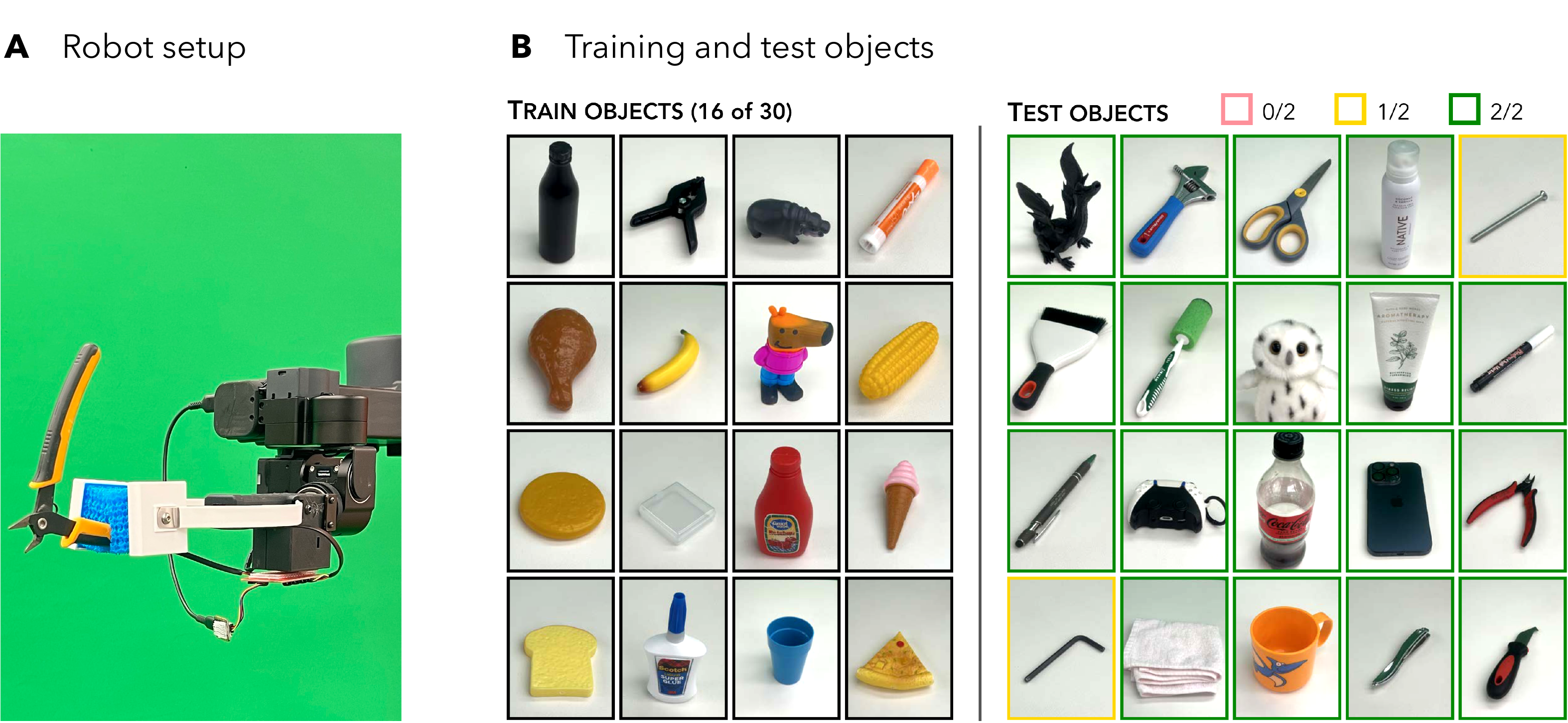}
    \caption{
    \textbf{Slip Detection.}
    (\textbf{A}) Hello Robot Stretch setup used for slip data collection and classifier evaluation.
    (\textbf{B}) Object sets: (Left) A diverse ensemble used for training and (Right) a held-out validation set of objects to evaluate our model on.
    }
    \label{fig:fig-4}
\end{figure}

A closely related and equally important capability for robots operating in unstructured environments is the ability to detect when an object has slipped from their grasp. In this experiment, we demonstrate that \name{} can be used to reliably detect object slip using a simple, generalizable linear classifier trained on interactions with a small set of objects. Our setup consists of a Hello Robot Stretch with integrated \name{} as shown in Fig.~\ref{fig:fig-4}A. A human operator slowly tugs on the grasped object for 1-2 seconds, after which a human annotator labels the sequence as “force” or “no-force” based on corresponding videos. We estimate three statistics from a sliding temporal window of the raw magnetometer signals: (1) the norm of the $x$ and $y$ components of the magnetic field corresponding to each of the five magnetometers, (2) the maximum change in sensor signals within the window, and (3) the standard deviation of the signal in the window, and use these statistics to train a linear binary classifier. The training dataset consists of four trajectories each of 30 everyday objects varying in shape, size, weight, and surface texture. To evaluate generalization, we test the model on a separate set of 20 unseen objects (Fig.~\ref{fig:fig-4}B). Despite its simplicity, the classifier achieves a high classification accuracy of 95\% on this held-out set.

\subsection{Visuotactile Policy Learning}
\label{subsec:policy-learning}
\begin{figure}[h]
    \vskip -0.5em
    \centering
    \includegraphics[width=\linewidth]{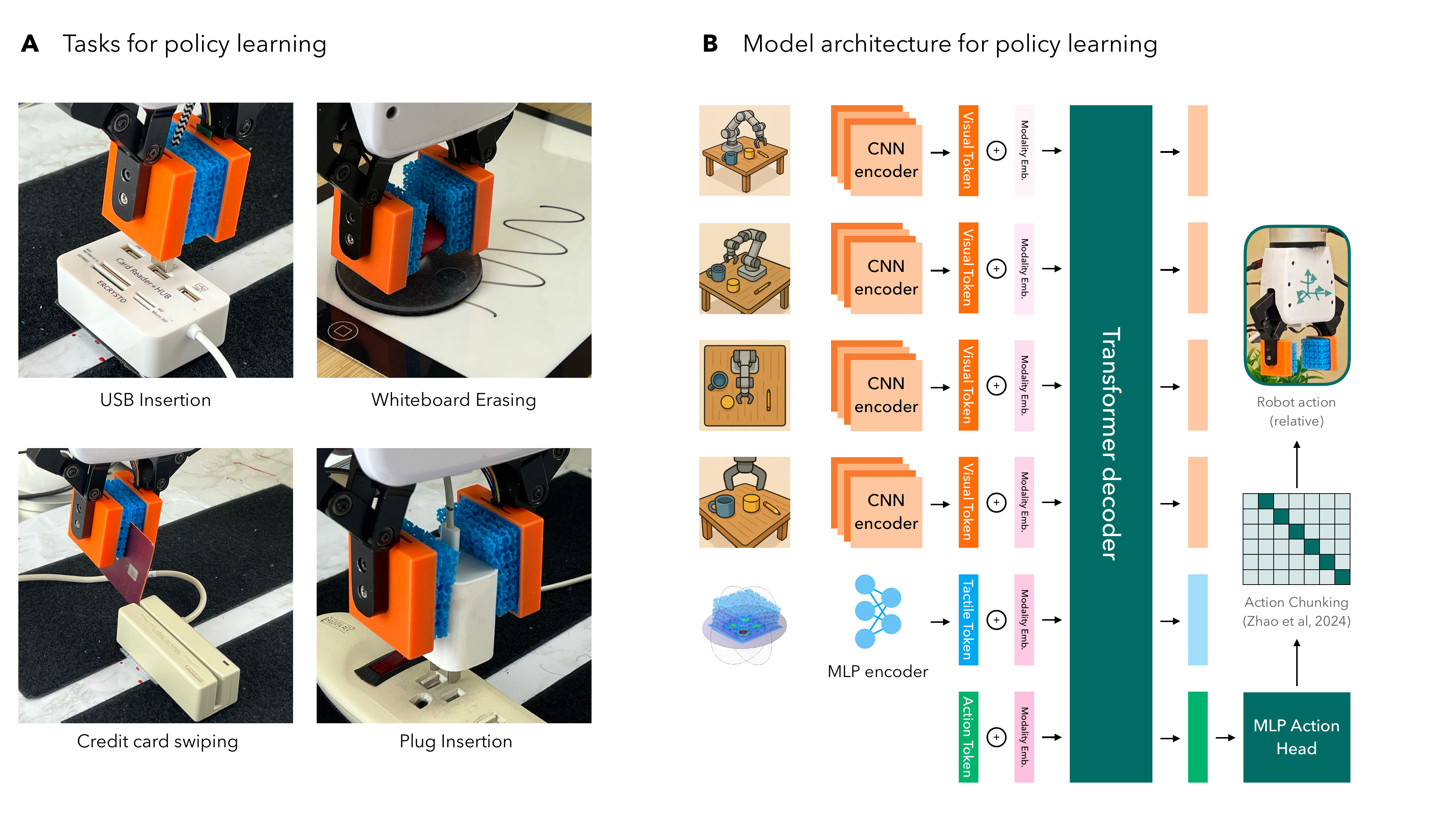}
    \caption{
    \textbf{Robot Policy Learning Tasks and Model Architecture.}
    (\textbf{A}) We evaluate the performance of \name{} on four precise tasks - USB Insertion, Whiteboard Erasing, Credit card swiping and Plug Insertion.
    (\textbf{B}) Visuo-Skin model architecture, used for our robot policy learning experiments. Images from the four cameras are encoded by a CNN encoder while the \name{} signal is encoded by an MLP encoder. The encoded features are passed through a transformer decoder policy along with an action token, whose output is used to predict an action chunk. This action chunk is supervised using demonstration data, and policies employ  exponential smoothing during deployment to avoid jerky motions~\cite{zhao2023learning}.
    }
    \label{fig:fig-5}
\end{figure}

Finally, a key application area for tactile sensors is in learning force-aware policies for robots for precise, contact-rich manipulation. We study the effectiveness of \name{} in visuotactile robot policy learning through four contact-rich manipulation tasks shown in Fig.~\ref{fig:fig-5}A. Drawing from the VisuoSkin~\cite{pattabiraman2024learning} framework, we employ a transformer-based architecture for learned policies using behavior cloning (Fig.~\ref{fig:fig-5}B). Here, we describe the robot environment, set of tasks, and the evaluation protocol.

\subsubsection{Environment Setup}
We use a Ufactory xArm 7 robot with its standard two-fingered gripper and an \name{} sensor attached to either fingertip as shown in Fig.~\ref{fig:fig-5}A. We use a Meta Quest 3 along with the OpenTeach~\cite{iyer2024open} teleoperation framework to collect demonstrations for behavior cloning. To improve dataset diversity and policy robustness, we adopt a strategy similar to VisuoSkin~\cite{pattabiraman2024learning}, introducing uniformly sampled angular perturbations to the commanded robot velocity at each timestep during teleoperation. We record visual data from four camera views: three third-person perspectives and one egocentric wrist-mounted camera attached to the robot’s gripper, all captured at 30 Hz. Simultaneously, we collect tactile data from the \name{} sensors in the form of raw magnetometer signals sampled at 100 Hz. All data streams are synchronized and resampled to 10 Hz for downstream training and deployment.

\subsubsection{Robot Task Descriptions}
We evaluate our approach on four precise, contact-rich manipulation tasks, drawing largely from AnySkin~\cite{bhirangi2024anyskin}, with an additional whiteboard erasing task. We choose tasks that require force-aware control and high precision for successful completion. In each task, the robot begins with a grasped object and must interact with a target object, while implicitly reasoning about contact forces (Fig.~\ref{fig:fig-5}A). To improve the robustness of the learned policies, we introduce variability by regrasping and slightly varying the orientation of the grasped object between demonstrations and evaluations. For each task, we conduct 30 evaluation trials, comparing the performance of our visuotactile policy against a vision-only baseline.

\textit{USB Insertion}:
In this task, the robot is initialized while grasping a USB charging cable, and the goal is to insert the cable into a designated USB socket mounted on the table. The training dataset for this task comprises 36 teleoperated demonstrations.

\textit{Plug Insertion}:
In this task, the robot is required to insert a plug into the first socket on a power strip fixed on a table. The training dataset consists of 48 demonstrations.

\textit{White-Board Erasing}:
The robot is initialized, with a whiteboard eraser grasped while a random region of a fixed white-board is scribbled on. The robot is required to locate the region, make firm contact, and erase the mark. We vary the region and size of the mark, keeping the location of the whiteboard fixed. The training dataset consists of 24 demonstrations.

\textit{Card swiping}:
In this task, the arm starts with a credit card grasped. The task is to locate the card reader and swipe the card through the reader. The training dataset consists of 32 demonstrations.

\subsubsection{Effect of \name{} on performance}
We observe that the trained visuotactile policies are able to successfully complete the aforementioned precise tasks, at a high average success rate of 91\%. Given the low margins of error needed for these tasks, this underscores both the signal consistency and the ability of \name{} to reliably capture the subtle sensory cues to then learn meaningful seeking behaviors in order to complete the tasks. Below, we break down the impact of \name{} for each of the tasks. 

\begin{table}[tbp]
    \centering
    \caption{
    \textbf{Learned robot policy performance}
    Average success rate (out of 10) over three evaluation trials of 10 rollouts each for learned, single-task policies on four precise tasks.}
    \label{tab:performance}
    \begin{tabular}{|cc|cccc|}
        \hline
        \multicolumn{2}{|c|}{\textbf{Input Modalities}} & \multicolumn{4}{c|}{\textbf{Success rate (out of 10)}} \\ \hline
        Vision & \name{} & USB Insertion & Plug Insertion & Whiteboard Erasing & Card Swiping \\
        \hline
        \textcolor{olivegreen}{Yes} & \textcolor{red}{No} & $3.33$ & $4.67$ & $5.67$ & $6.33$ \\
        \textcolor{olivegreen}{Yes} & \textcolor{olivegreen}{Yes} & \textbf{8.33} & \textbf{9.33} & \textbf{9.67} & \textbf{9.00} \\
        \hline
    \end{tabular}
\end{table}

For the USB insertion and plug insertion tasks, while the vision-only policies occasionally insert into the designated slot, they exhibit a headfirst behavior by pushing down at the first point of contact with the target object. With the addition of \name{}, the policy learns a seeking behavior guided by contact force to confirm alignment with the target slot before pushing down to insert. The most common mode of failure for the learned visuotactile policy for plug insertion is mistaken alignment, where the policy pushes downward when the two-pronged plug makes contact with the ground pin slot. In the case of USB insertion, this mode is orientation mismatch, where the USB cable is slightly misaligned before insertion, resulting in failure. We see similar corrective and seeking behaviors for the visuotactile policy in the card swiping task. Vision-only baselines exhibit two modes of failure in this task -- failing to distinguish between full vertical insertion into the card reader, and being unable to recover from a collision at the point of entry, often bending the card outward and rendering it unable to complete the swipe successfully.

In the whiteboard erasing task, while both visual and visuotactile policies are able to reach the region of interest correctly, the visual policy is unable to exert the right amount of downward force on the whiteboard. It either begins executing the erasing motion before making contact or continues pushing downwards even after making contact. With the addition of \name{} sensors, the robot moves downward until it makes firm contact with the whiteboard and then immediately begins swiping and completes the task.

\subsection{Sensor properties and customization}
\label{subsec:sensor-properties}
\begin{figure}[h]
    \vskip -0.5em
    \centering
    \includegraphics[width=\linewidth]{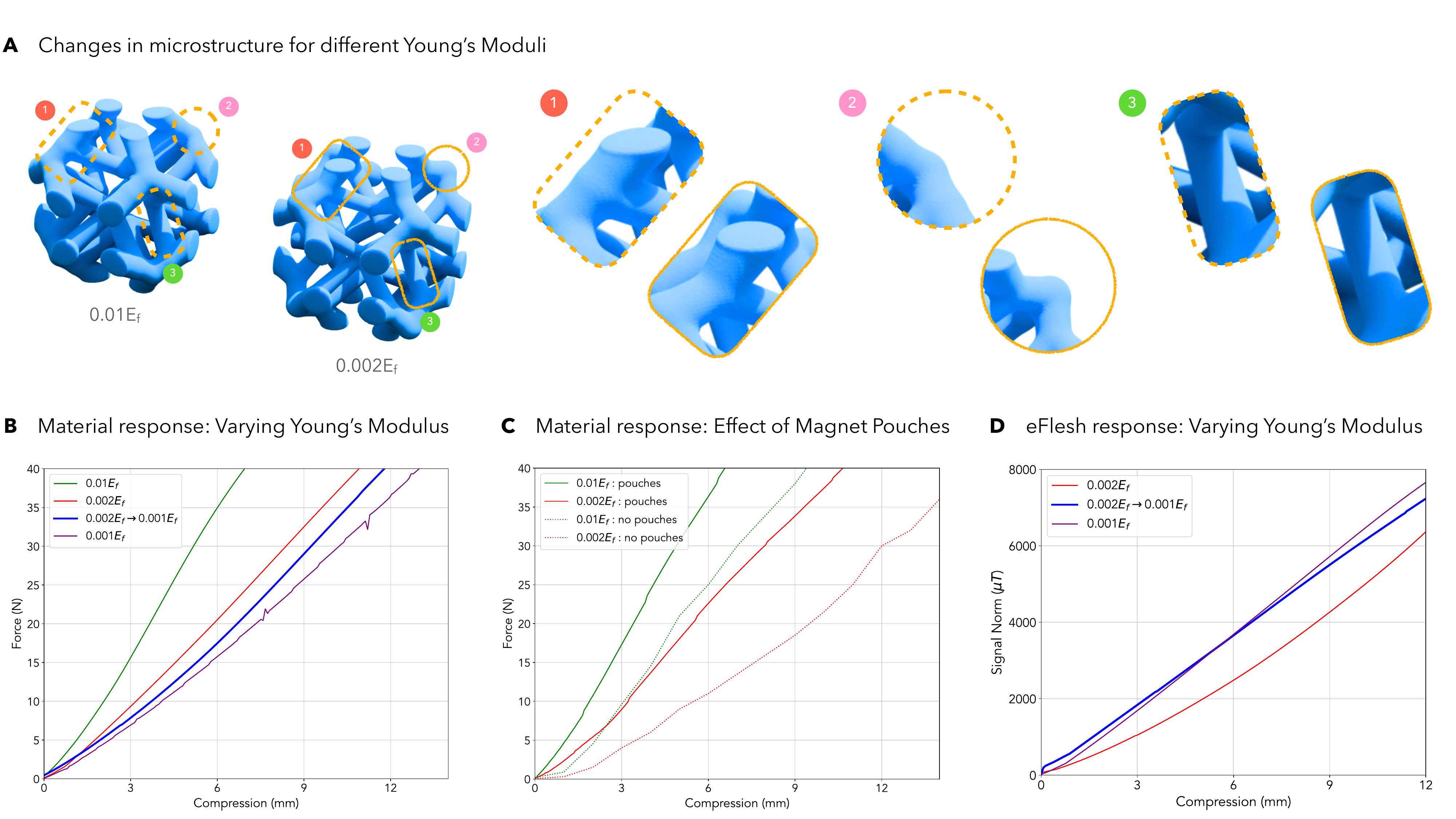}
    \caption{
    \textbf{Effect of Varying Young's Moduli on Sensor and Material Response.}
    (\textbf{A}) Microstructure variants with two different specified Young's Moduli - $0.01E_f$ and $0.002E_f$; the lower modulus microstructure exhibits correspondingly thinner beams.
    (\textbf{B}) Force-Compression curves for different variants of \name{}, with different specified Young's Moduli; higher modulus corresponds to stiffer material response.
    (\textbf{C}) Effect of magnet pouches on the material response; with the addition of pouches and magnets, the material response is more rigid, but the trends with respect to specified Young's Modulus are consistent.
    (\textbf{D}) Signal Norm-Compression curves; depth-graded stiffness of high ($0.002E_f$) between magnets and \name{} surface, to low ($0.001E_f$) between magnets and the Hall sensor PCB, produces higher signal at low-medium compression. This is because the region beneath the magnets compresses more, for the same surface displacement.
    }
    \label{fig:fig-6}
\end{figure}

In addition to the geometric customizability outlined before, the microstructure-based design also enables modulation of \name{} stiffness, and therefore, the resulting response characteristics. Building on the framework proposed by Tozoni et al.~\cite{Tozoni2024}, we adjust the mechanical properties of the sensor by tuning the parameters of its repeating unit cells, mainly the beam dimensions and the cubic cell size. For fixed beam dimensions, larger cell size results in lower stiffness. Conversely, for a fixed cell size, larger beam dimensions result in higher stiffness. Intuitively, smaller and denser microstructures produce stiffer sensors with lower sensitivity but a wider operating range, while larger, more compliant cells increase sensitivity at the cost of operating range. While the exact relationship between microstructure parameters and material response is beyond the scope of this paper, Fig.~\ref{fig:fig-6}A qualitatively demonstrates the changes in microstructure as the desired stiffness is varied. Due to print resolution constraints of the 3D printer used in our fabrication, we limit minimum beam size to 0.4 mm and above, but the framework naturally extends to finer geometries with higher-resolution 3D printers.

To enable users to tailor sensor response to their application needs, we characterize multiple variants of \name{} sensors parameterized by target Young's Modulus values defined relative to the Young's Modulus of the filament, $E_f$. For this experiment, we use the same Ufactory xArm 7 setup as previous experiments equipped with a flat indenter (Fig.~\ref{fig:fig-2}B) and place the sensor on a weighing scale. We apply controlled surface deformations in increments of $0.1$ mm on its surface using the robot, recording the force measured by the scale and the raw magnetometer signals at each step. This allows us to generate force-displacement curves for each sensor as shown in Fig.~\ref{fig:fig-6}B. Further, this experiment also captures the force sensitivity of each sensor variant. Here, we define the force sensitivity as the minimum force when the norm of the raw signal is $6 * \sigma_\text{noise}$, where $\sigma_\text{noise}$ is the standard deviation of the sensor response when no force is applied. We report a sensitivity of 0.04 N for the depth-graded \name{} variant ($E = 0.002E_f \rightarrow 0.001E_f$ from top layer to bottom) and 0.23 N for the stiffer \name{} variant with uniform $E = 0.01E_f$. Notably, even the stiffest \name{} variant exhibits a higher sensitivity in comparison to AnySkin, which has a sensitivty of 0.64 N.

While the microstructure optimization method by Tozoni et al.~\cite{Tozoni2024} was originally designed for small-strain regimes, \name{} sensors operate in large-strain regimes and contain inhomogeneities introduced by embedded magnet pouches. Despite this, we observe that the overall trends hold -- i.e., higher Young’s modulus corresponds to stiffer response (Fig.~\ref{fig:fig-6}B). However, to quantify the impact of embedded magnets on mechanical response, we compare force-displacement curves of two sensorized blocks with their unsensorized counterparts, fabricated with identical parameters. We find that the addition of magnets increases overall stiffness (Fig.~\ref{fig:fig-6}C), a factor users should consider during design. Nonetheless, for simplicity and intuitiveness, we retain Young’s modulus as the primary design specification parameter in our open-source slicing tool.

Finally, this design flexibility allows for spatial variation of stiffness within the sensor. Since \name{} transduces surface deformation through magnet displacement, amplifying magnet motion can enhance signal strength. In the experiments presented here, we use a graded stiffness configuration, with lower stiffness (e.g., $E = 0.001E_f$) beneath the magnets to allow greater displacement for a given surface deformation, and higher stiffness above ($E = 0.002E_f$) to maintain structural support. As shown in Fig.~\ref{fig:fig-6}D, this configuration results in stronger signal responses than a uniformly stiff sensor, without compromising mechanical durability.

\subsubsection{Effect of magnetic interference}
\label{subsec:interference}
\begin{figure}[h]
    \vskip -0.5em
    \centering
    \includegraphics[width=\linewidth]{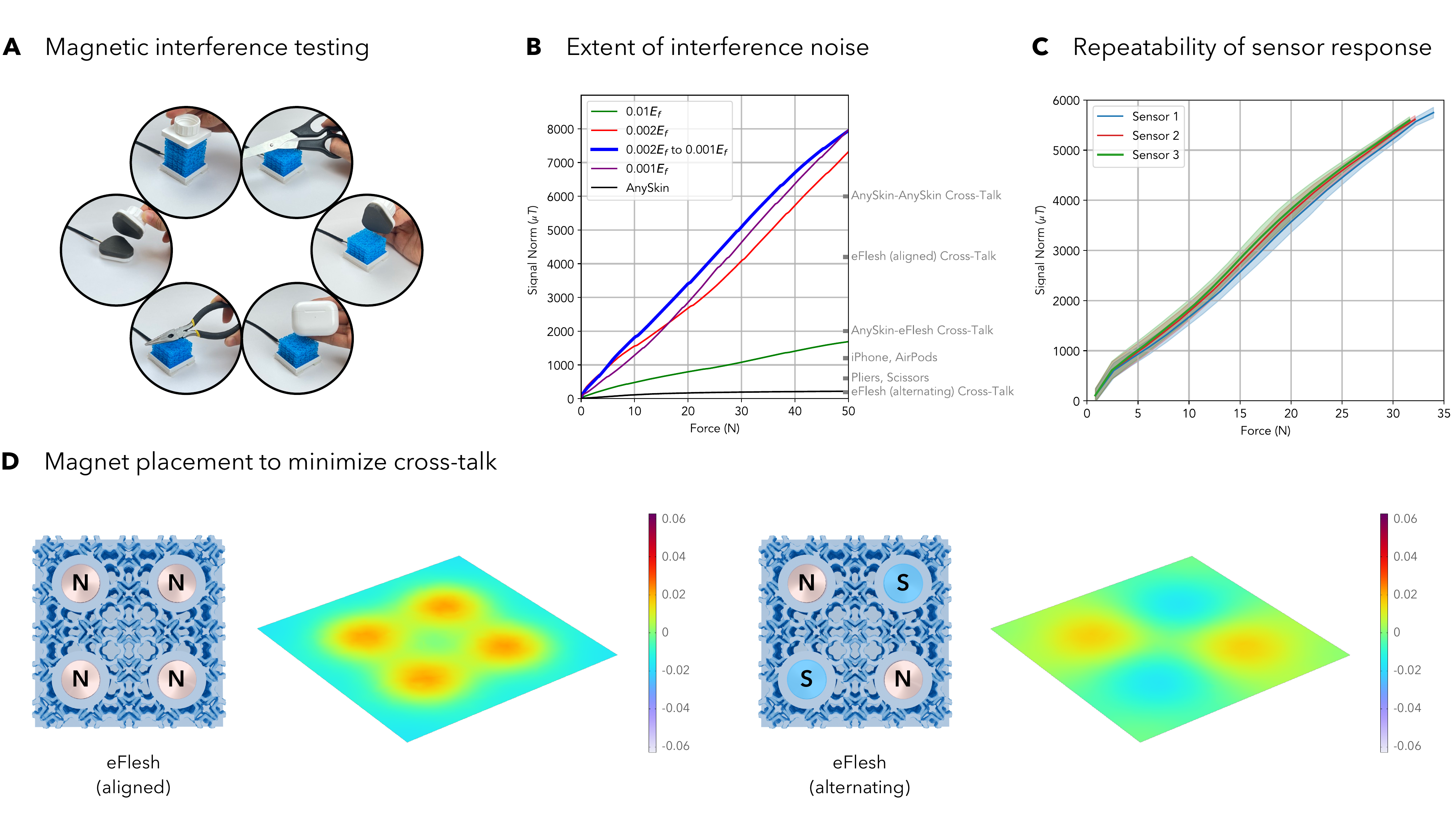}
    \caption{
    \textbf{Magnetic Interference Mitigation and Repeatability.}
    (\textbf{A}) Experimental set-up for magnetic–interference testing.  Everyday ferromagnetic or magnetized objects (iPhone, AirPods case, pliers, scissors, etc.) are brought into close proximity with an \name{} instance and the maximum signal norm is recorded.
    (\textbf{B}) Signal norm - Force curves for \name{} variants and quantification of interference from magnetic objects and other sensor instances. \name{} cross-talk corresponds to a sensor signal from less than 0.2 N of force.
    (\textbf{C}) Repeatability of \name{} sensor response.  Three independently fabricated \name{} sensor instances were cycled through incremental normal loads; mean response (solid line) and one-standard-deviation band (shaded) show less than $5\%$ coefficient of variation over the force range.
    (\textbf{D}) Fabrication strategy to minimize \name{}-\name{} cross-talk. Alternating magnet pattern (right) cancels the dominant far-field components, yielding a reduction of two orders of magnitude in stray-field intensity and, consequently, cross-talk compared to the sensor instance with aligned magnet polarities (left). Heatmaps depict the z-component of the magnetic flux density (in T) measured on a plane 15mm above the magnet array, i.e., at the surface of the cuboidal \name{}.
    }
    \label{fig:fig-7}
\end{figure}

A critical limitation of existing magnetic tactile sensors~\cite{tomo2018new,bhirangi2021reskin,bhirangi2024anyskin} is their susceptibility to interference from nearby ferromagnetic materials and electromagnetic devices, such as smartphones and other electronics. To better understand the relative magnitude of magnetic interference compared to tactile signals from deformation, we move a few household objects around all sides of the sensor as shown in Fig.~\ref{fig:fig-7}A, and record the maximum norm of the resulting noise recorded by the Hall effect sensors. We then use the flat indenter experimental setup of the previous section, and compare it against these signals, as shown in Fig.~\ref{fig:fig-7}B. This experiment demonstrates that the objects with the strongest magnetic field correspond to a signal distortion of less than 1mm surface deformation. 

Beyond everyday objects, the most conspicuous source of magnetic interference is the presence of other \name{} sensors in the environment. As shown in Fig.~\ref{fig:fig-7}B, the magnetic field produced by a neighboring sensor (indicated by eFlesh(aligned) cross-talk) can induce a signal magnitude comparable to that generated by substantial surface deformation -- in some cases, up to half the sensor’s maximum deformation range. This level of interference poses a serious challenge for applications involving multi-sensor setups, such as dexterous manipulation, where sensors must be placed in close proximity. To address this issue, we introduce a simple yet effective modification to the \name{} design: the embedded magnets are arranged in \textit{alternating} instead of \textit{aligned} polarities. This configuration, as demonstrated in Fig.~\ref{fig:fig-7}D, significantly attenuates the strength of the magnetic field outside the sensor, minimizing cross-sensor interference, which now corresponds to negligible sensor deformation. 

This stands in stark contrast to prior sensors such as AnySkin~\cite{bhirangi2024anyskin}, which exhibit substantial magnetic interference relative to their signal strength. Analyzing the signal from AnySkin indicates that the magnitude of the largest AnySkin cross-talk signal corresponds to more than 20$\times$ the maximum deformation signal from AnySkin. On the other hand, the maximum cross-talk signal for \name{} corresponds to an applied force of less than 0.2N, clearly demonstrating a drastic improvement in robustness to magnetic interference.

\subsubsection{Cross-instance consistency}
As tactile sensors seek widespread adoption, consistency across sensor instances becomes a critical requirement, often unaddressed in past work~\cite{yuan2017gelsight,lambeta2020digit}. While the deformable nature of soft sensors enables stable, conformal contact with diverse surfaces, it also makes them susceptible to wear and damage from repeated compression cycles -- necessitating periodic replacement. In this context, cross-instance signal compatibility is essential not only for ensuring plug-and-play usability, but also for enabling data reusability in training data-driven models, particularly relevant if tactile sensors are to match the scale and generality of large vision and language models. Collecting diverse, high-quality datasets is only valuable if the underlying signals are consistent across sensors, allowing data from different sensor instances to be aggregated seamlessly.

To evaluate signal consistency across sensor instances, we characterize the signal variation by measuring the norm of the \name{} signal as a function of applied normal force, as shown in Fig.~\ref{fig:fig-7}C. Across the full deformation range, we observe a maximum standard deviation of a signal corresponding to less than 1N between different instances, indicating good alignment in raw signal response. 

%% file: sections/3_discussion.tex
\section{Discussion}

In this work, we introduced \name{}, a 3D-printable, magnetic sensor that offers a low-cost and accessible approach to tactile sensing. By leveraging only a standard 3D printer, a small number of embedded magnets, and a magnetometer circuit board, \name{} enables sensorizing a wide range of robots. This design opens the door to democratizing the fabrication of tactile sensors, significantly improving accessibility and customizability for in-the-loop design iteration.

The broader significance of \name{} lies in its potential to equip robots with richer sensing, particularly as we deploy them in unstructured, dynamic environments. Vision alone is often insufficient in a number of tasks requiring fine-grained spatial awareness, especially in occluded and cluttered settings. \name{} provides a lightweight, conformal, and minimally disruptive tactile solution with high sensitivity and a large dynamic range, in addition to shape and response customizability.

A key advantage of \name{} over traditional magnet-based sensors~\cite{tomo2018new} and existing soft sensing techniques~\cite{bhirangi2024anyskin,huang3d,Xu2024CushSenseSS} is its ease of fabrication and scalability. Unlike previous systems that rely on manual  assembly, specialized equipment, or complex, multi-step fabrication, \name{} is entirely 3D-printable and modular. This lowers the barrier to entry as well as prototyping time for users.

However, magnetic sensing in uncontrolled environments remains a challenge. One known limitation is susceptibility to magnetic interference, especially in proximity to electromagnetic or ferromagnetic objects, including other instances of the magnetic sensor itself. While \name{} effectively mitigates interference from most everyday sources and nearby \name{} sensors, it may still encounter stronger magnetic fields in industrial settings such as power infrastructure or server environments. In future work, we aim to investigate materials with high magnetic permeability and explore passive shielding techniques to further reduce these effects.

Another open question concerns scaling and generalization. Our current results show strong promise in small-scale experiments, as well as signal consistency across instances conspicuously absent across the spectrum of existing tactile sensors. This data reusability along with robustness to magnetic interference makes \name{} uniquely amenable to data collection and deployment at scale across a diverse set of robotic platforms and environments. To this end, future work will involve large-scale data collection in realistic settings along the lines of prior work in home robotics~\cite{shafiullah2023bringing, etukuru2024robot}, to validate generalization and robustness. We also envision learning tactile representations from these datasets to extract richer features from magnetic signatures, further enhancing applicability in novel scenarios.

Finally, the modular nature of eFlesh suggests exciting possibilities for integration into multi-functional systems combining sensing and actuation. By allowing sensing capabilities embedded directly into the structure of robots, \name{} represents a step forward in material-based robotics that closely integrates form and function in the construction of robots.

%% file: sections/4_materials.tex
\section{Materials and Methods}

\subsection{Fabrication of \name{}}

The cuboidal instance of \name{} used in all of the experiments presented above is composed of three layers of 5 $\times$ 5 microstructure grids stacked on top of each other. The Young's modulus of each layer is constant across the grid plane and increases linearly from the bottom layer to the top layer as $0.001E_f$, $0.0015E_f$ and $0.002E_f$ respectively, where $E_f$ represents the Young's modulus of the filament (we use TPU 95A). Each microstructure cell is a cube of size 8mm, resulting in overall sensor dimensions of 40mm $\times$ 40mm $\times$ 24mm. After generating the microstructure grid, we add four lip-sealed pouches in the middle layer, distributed as shown in Fig~\ref{fig:fig-1}. These pouches are designed to press-fit tolerance for N52 neodymium magnets that each have a diameter of 9.525mm and a thickness of 3.175mm.

This workflow is adaptable to any arbitrary 3D geometry, specified as a \texttt{.obj/.stl} file. Once the Young's modulus is specified, cell size is primarily constrained by the minimum printable thickness of the beams that make up each cell. On deciding a cell size, the number of layers can be computed by dividing the object's height by this cell size. We first create a cuboidal block of dimensions of the input mesh, thus encompassing it, composed of cut-cells of the specified cell size and Young's Modulus. Next, we trim the region that extends beyond the convex hull of the input shape, to preserve the surface shape. The user can then specify their cylindrical magnet dimensions and position, following which we parametrically add lip-sealed pouches within the microstructure grid.

We fabricate the sensors using a Bambu X1 Carbon 3D printer with a standard 0.4mm nozzle, printing with TPU filament with a Shore hardness of 95A, requiring no print supports. The model is processed in OrcaSlicer, an open-source slicing tool, which is configured to pause the print one layer before the magnetic pouch covers. At this point in the print, the user inserts magnets into the pouch packets, with the north poles facing upwards in this case, but in general, polarities may be altered based on the considerations outlined in the previous sections. Printing is then resumed to encase the magnets in the pouch as well as complete the rest of the sensor structure.




\subsection{Neural Architectures and Training}
In this section, we provide an overview of the training details and model architectures used for the sensor characterization and robot policy learning experiments. 

\subsubsection{Spatial and Force Resolution}
We use a multi-layered perceptron (MLP) with two hidden layers, each with 128 nodes and ReLU activations. The network is trained using mean squared error (MSE) loss, with the Adam optimizer, a learning rate of \texttt{1e-3} and a batch size of 64. Target outputs - $x,y,z$ coordinates are normalized for training using their corresponding ranges of 30 mm for $x$ and $y$ and 4 mm for $z$. RMSEs are reported after unnormalizing the predictions. Training is performed on a single NVIDIA RTX 3080 GPU over a maximum of 1000 epochs, requiring under 5 GPU minutes.

\subsubsection{Visuo-Tactile Robot Learning}
Drawing from Visuo-Skin~\cite{pattabiraman2024learning}, we use a multi-sensory transformer architecture~\cite{haldar2024baku}, where we fuse the visual and tactile observations as a modified ResNet-18~\cite{he2016deep} encoding and a 2-layer MLP encoding, respectively, projected to the same dimensionality. 

Visual data is resized to 128x128 and the \name{} data corresponds to 15-dimensional vectors for each fingertip, and the time-synchronized visuo-tactile dataset is subsampled by a factor of 5. We tokenize each of the two fingertips' \name{} data separately prior to feeding them into the transformer decoder. Robot policies are then trained on the subsampled dataset up to 36,000 checkpoints, requiring 3 hours on a single NVIDIA RTX 8000 GPU.

%% file: sections/6_appendix.tex
\section{Supplementary Materials}

\subsection{CAD-to-eFlesh Conversion Tool}

\textbf{Prerequisites:} A CAD file (.obj/.stl) of the part to sensorize, an installation of OrcaSlicer and optionally an installation of Blender v4.2.

\noindent\textbf{Stage 1:} Generate microstructure version of input shape

\begin{figure}[h]
    \vskip -0.5em
    \centering
    \includegraphics[width=0.35\linewidth]{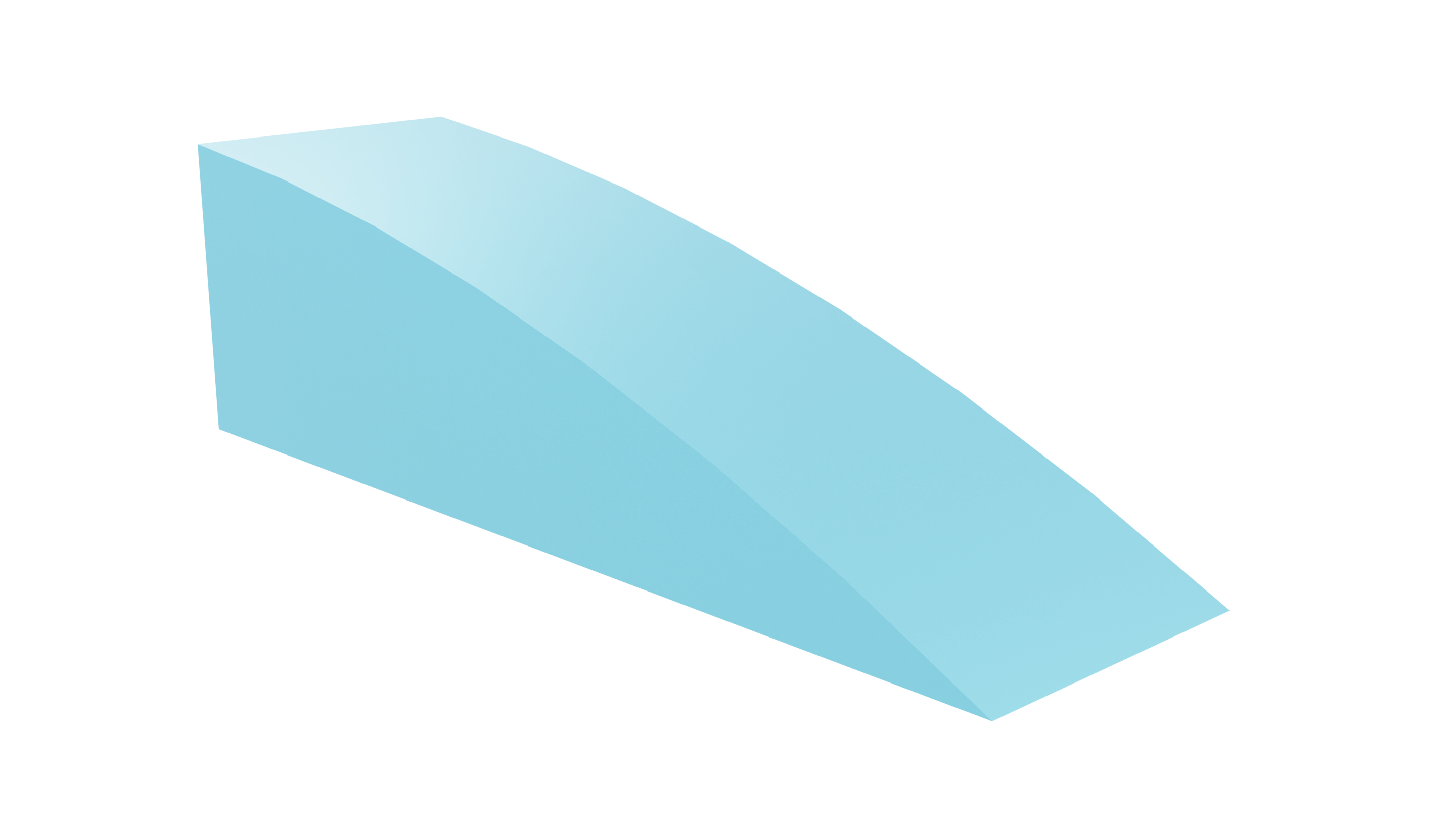}
    \caption{
    \textbf{Input CAD file to sensorize},
    A gripper finger fed as input to our conversion tool.
    }
    \label{fig:umi-input}
\end{figure}

\begin{figure}[ht]
    \vskip -0.5em
    \centering
    \begin{minipage}[t]{0.35\linewidth}
        \centering
        \includegraphics[width=\linewidth]{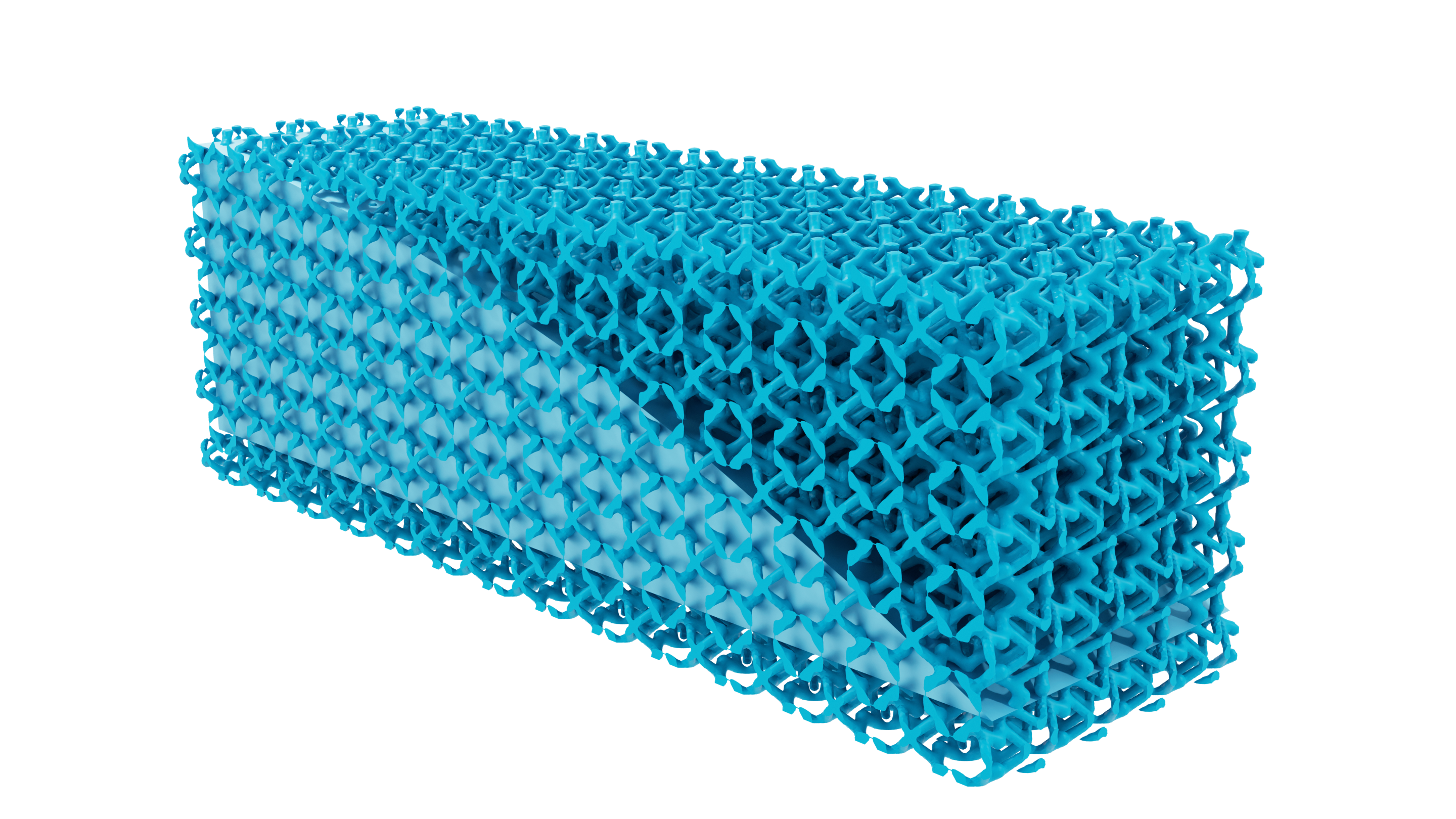}
        \caption*{
        (\textbf{A}) 
        }
    \end{minipage}
    \hfill
    \begin{minipage}[t]{0.35\linewidth}
        \centering
        \includegraphics[width=\linewidth]{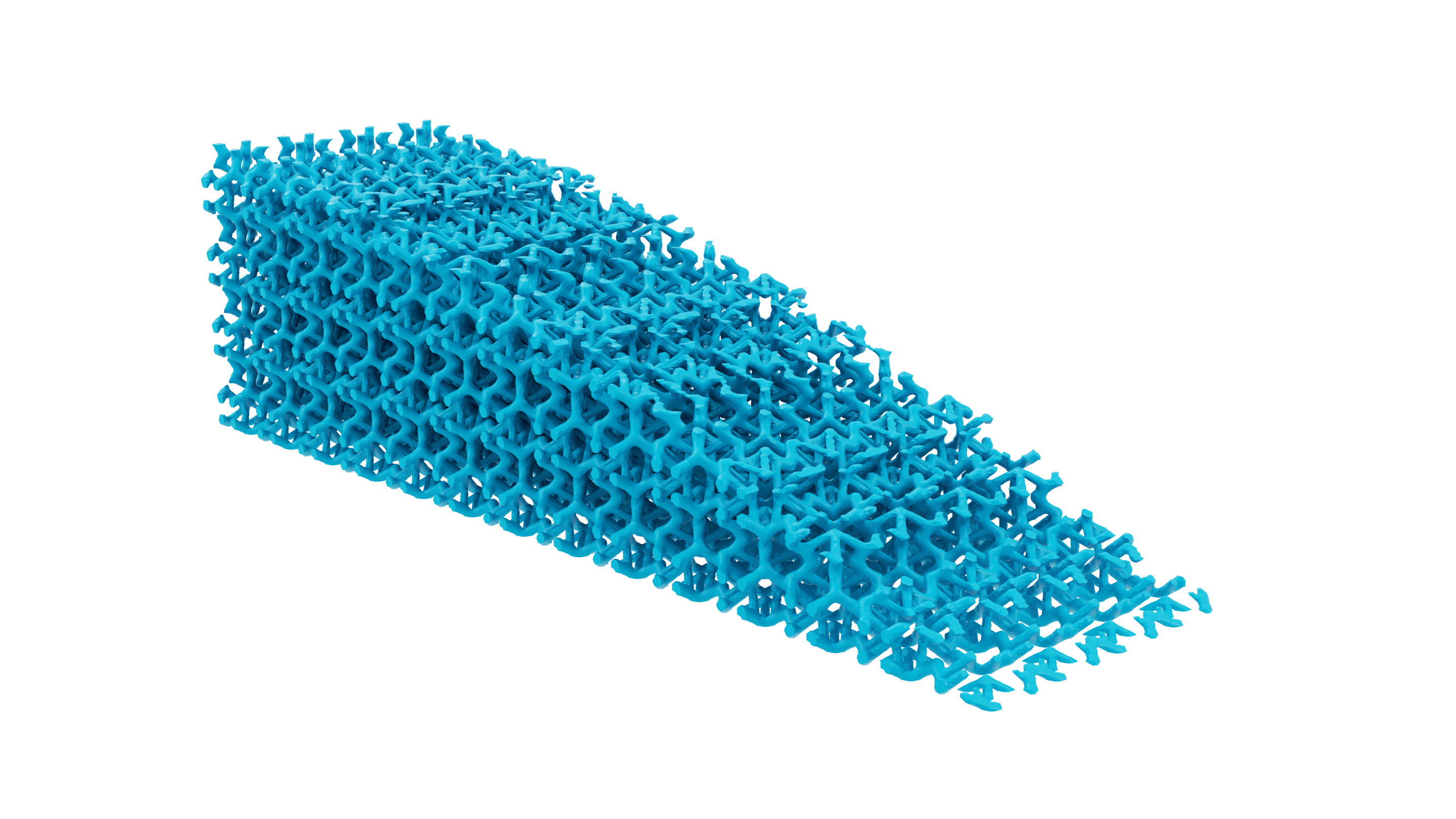}
        \caption*{
        (\textbf{B}) 
        }
    \end{minipage}
    \caption{
    \textbf{Illustration of the trimming process, to generate the microstructure version of the input CAD file}
    (\textbf{A}),
    The input shape is first fully encompassed by a cuboidal microstructure grid composed of repeating cut-cell microstructure cells.
    (\textbf{B}), 
    After the trimming boolean operation, the microstructure retains the outline and shape of the input gripper finger.
    }
    \label{fig:umi-lattice}
\end{figure}

\noindent The first stage involves generating a microstructure grid and trimming it to fit the input shape (Figure~\ref{fig:umi-input}). A cuboidal grid of dimensions as the length, breadth, and height of the input shape is first generated (Figure~\ref{fig:umi-lattice}A), by repeating the composing microstructure units, characterized by the specified cell size and Young's modulus (Section~\ref{subsec:sensor-properties}). Following this, using the outline of the input shape, the grid region extending outwards from its convex hull is trimmed. The trimming is done by performing a Boolean intersection with an infinitely large object that has a hole matching the shape of the convex hull--effectively removing everything outside it (Figure~\ref{fig:umi-lattice}B).

\noindent Input: CAD file, cell size, Young’s modulus (optionally spatially varied by layer) represented as a factor of the Young's modulus of the composing filament ($E_f$).

\noindent Output: Latticed OBJ file fitted to the shape of the input CAD file.

\noindent\textbf{Stage 2:} Add pouches for magnets

\begin{figure}[h]
    \vskip -0.5em
    \centering
    \includegraphics[width=0.35\linewidth]{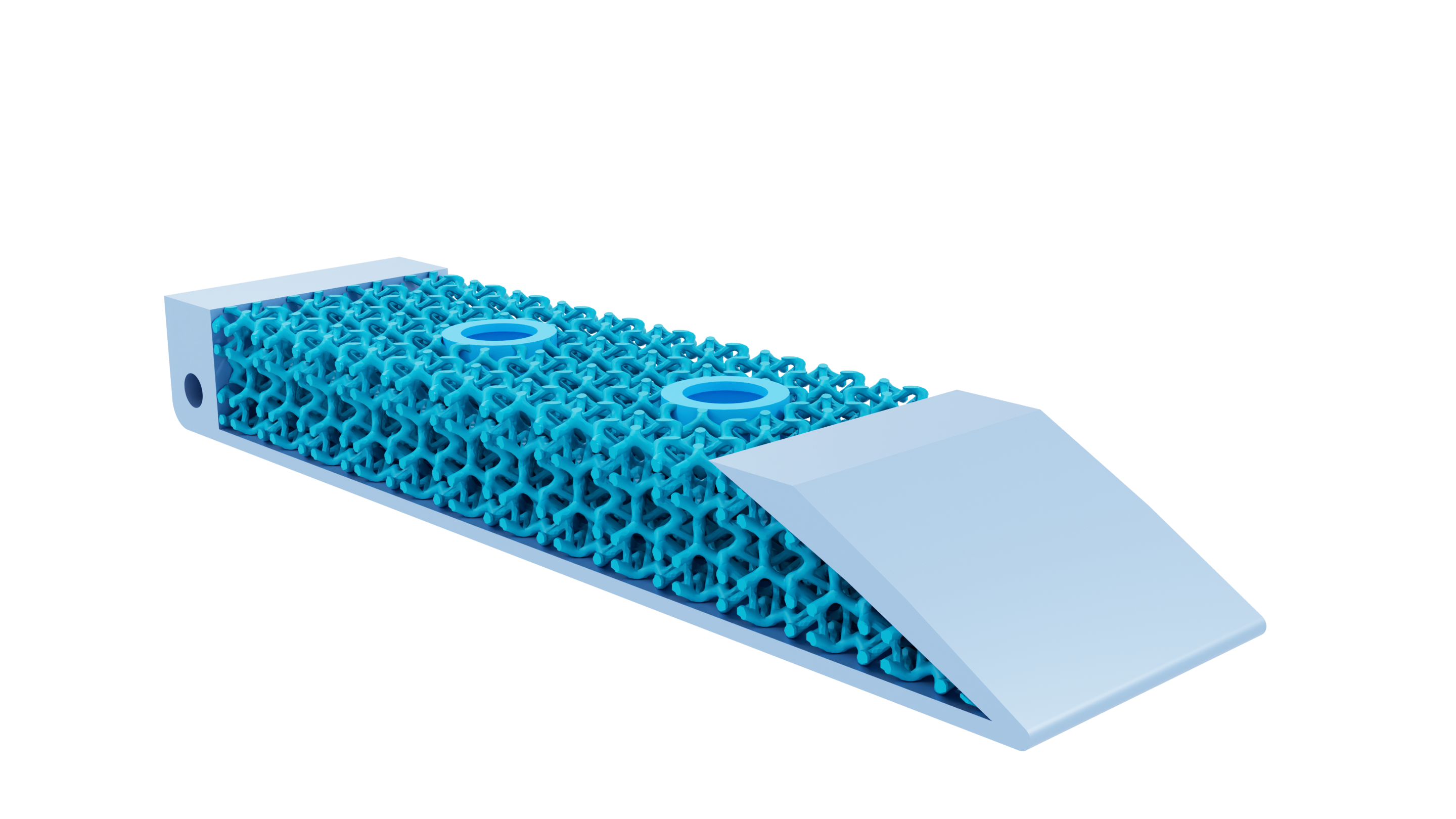}
    \caption{
    \textbf{Microstructure version with added pouches for magnets}
    }
    \label{fig:umi-pouches}
\end{figure}

\noindent The second stage involves placing an arbitrary number of pouches with specified magnet dimensions at selected locations (Figure~\ref{fig:umi-pouches}). Two pouch placement options are provided:

\begin{itemize}
    \item \textit{Option 1:} Use TinkerCAD, a browser-based application. We provide pouch files for magnets of two different dimensions— (i) diameter: 9.525mm, thickness: 3.175mm; and (ii) diameter: 4.763mm, thickness: 1.588mm. Users can position these pouches arbitrarily within their latticed file and then 'group' them to perform a Boolean addition.
    \item \textit{Option 2:} Users with Blender v4.2 installed can use our provided script. The script automatically performs Boolean operations to position pouches inside a given input model. The user can specify the number of pouches, the size and position of each pouch in the script.
\end{itemize}

\noindent\textbf{Stage 3:} Add slot(s) for the Hall sensor / magnetometer circuit board

\begin{figure}[h]
    \vskip -0.5em
    \centering
    \includegraphics[width=0.35\linewidth]{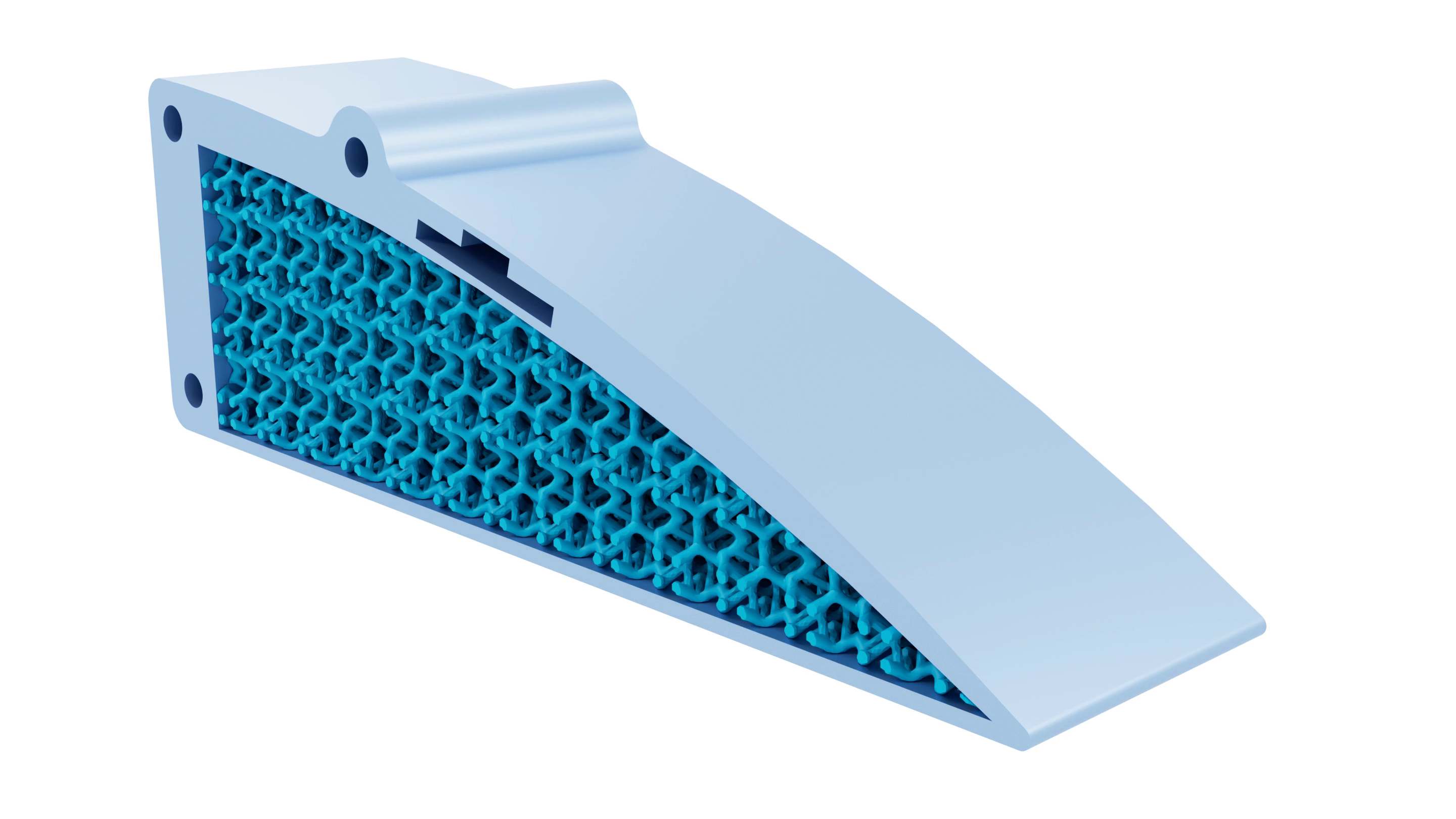}
    \caption{
    \textbf{Printable eFlesh},
    eFlesh with pouches and a slot for the magnetometer PCB.
    }
    \label{fig:umi-slot}
\end{figure}

\noindent Using any design software, the user creates a slot with appropriate dimensions for inserting the magnetometer. This region must not be within the lattice structure but instead be placed in a fully infilled region (if using TPU) or in a suitable infill region for rigid filaments like PLA or ABS (Figure~\ref{fig:umi-slot}).

\noindent\textbf{Stage 4:} Slice and 3D print the file

\begin{figure}[h]
    \vskip -0.5em
    \centering
    \includegraphics[width=0.45\linewidth]{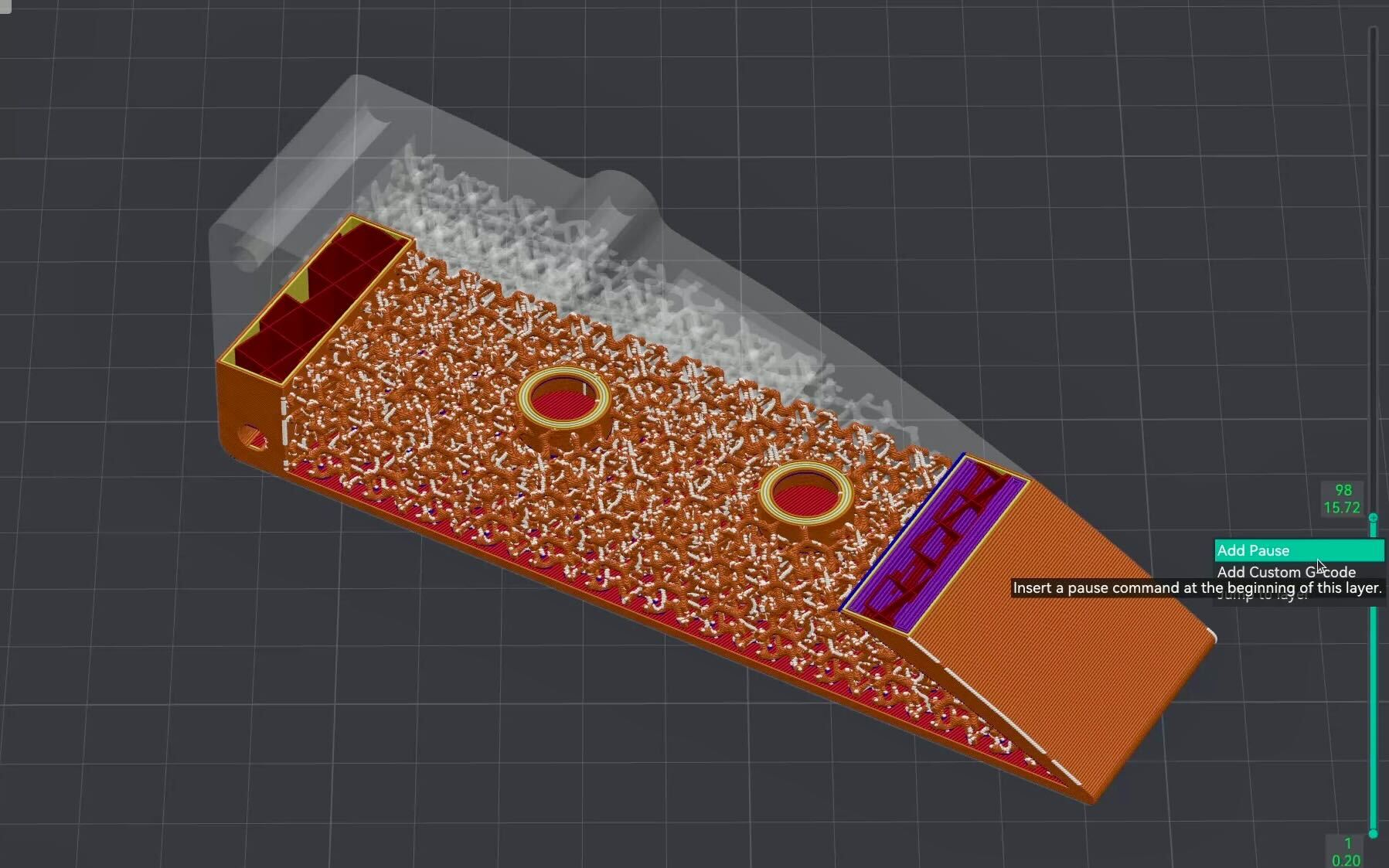}
    \caption{
    \textbf{eFlesh print sliced with a pre-programmed pause}, using the OrcaSlicer software.
    }
    \label{fig:umi-orca}
\end{figure}

\noindent Import and slice the generated file with TPU filament. Ensure magnet pouches are oriented parallel to the print bed. Users can preview the print layers after slicing, scroll to identify the layer at which the magnet pouches close. A pre-programmed pause can be inserted by right-clicking on the layer bar at this layer and clicking on 'Add Pause' (Figure~\ref{fig:umi-orca}). When the print pauses at the designated layer, the user inserts the magnets and resumes printing to complete the fabrication of the eFlesh sensor instance.

